\documentclass[conference]{IEEEtran}
\IEEEoverridecommandlockouts
\newif\iftechrep

\techrepfalse   % NORMAL PAPER
\usepackage{cite}
\usepackage{amsmath,amssymb,amsfonts}
\usepackage{algorithmic}
\usepackage{graphicx}
\usepackage{textcomp}
\usepackage{xcolor}
\usepackage{url}
\usepackage{tabularx}
\usepackage{booktabs}

\definecolor{azure}{RGB}{240,248,255} % soft readable azure

\usepackage[most]{tcolorbox}
\tcbset{
  keyfindingbox/.style={
    colback=gray!10, colframe=black!70,
    boxrule=0.5pt, arc=2pt, outer arc=2pt,
    left=6pt, right=6pt, top=4pt, bottom=4pt,
    fonttitle=\bfseries, before skip=10pt, after skip=10pt
  }
}

\def\BibTeX{{\rm B\kern-.05em{\sc i\kern-.025em b}\kern-.08em
    T\kern-.1667em\lower.7ex\hbox{E}\kern-.125emX}}

\usepackage{listings}

\begin{document}

\title{Large Language Models as Digital Twins:\\a Distribution-Aware Questionnaire-Based Evaluation\\
%{\footnotesize \textsuperscript{*}Note: Sub-titles are not captured in Xplore and
%should not be used}
%\thanks{Identify applicable funding agency here. If none, delete this.}
}

\author{
\IEEEauthorblockN{
    Francesco Di Cursi\IEEEauthorrefmark{1}\IEEEauthorrefmark{2},
    Chiara Boldrini\IEEEauthorrefmark{1},
    Andrea Passarella\IEEEauthorrefmark{1},
    Marco Conti\IEEEauthorrefmark{1}
}
\IEEEauthorblockA{
    \IEEEauthorrefmark{1}
    Institute of Informatics and Telematics (IIT),
    National Research Council (CNR), Pisa, Italy
}
\IEEEauthorblockA{
    \IEEEauthorrefmark{2}
    Department of Computer Science,
    University of Pisa, Pisa, Italy
}
\IEEEauthorblockA{
    \{francesco.dicursi, chiara.boldrini,
    andrea.passarella, marco.conti\}@iit.cnr.it
}
}

\maketitle

\begin{abstract}
This paper investigates whether Large Language Models can answer a personality questionnaire as a specific individual would. Given a user's social-media posts, the model completes the BFI-2-60 Big Five questionnaire as that user. The predicted responses are converted into trait scores and compared with the user's actual scores.

We evaluate GPT-4, Gemma, Llama 3.1, and Mistral on the MyPersonality dataset. To determine whether predictions reflect information from the user's writing, we compare them with answers produced without user text and with values derived from general population tendencies. We measure prediction error across all users and separately for individuals with low, intermediate, and high trait scores. We also test whether an LLM can express an explicit personality profile when given only the target Big Five scores, without user text, and examine how predictions change as more text is provided.

The results show that user text does not consistently improve performance. Models perform better near the middle of the trait distribution, while at the lower and upper ends they often remain close to central values or perform well at one end but poorly at the other. When given an explicit profile, GPT-4 most clearly translates trait information into consistent questionnaire responses, particularly for Extraversion and Neuroticism. Increasing context does not systematically reduce error: open-source models fluctuate more, whereas GPT-4 remains comparatively stable. Overall, the evaluated models do not consistently reproduce the questionnaire behavior of specific individuals.
\end{abstract}


\iftechrep
    % Technical-report mode:
    % skip the entire main paper and continue to the appendix.
\else

\section{Introduction}

Automatic Personality Prediction from Text (APPT) is a well-established task in NLP and machine learning. Motivated by the Lexical Hypothesis~\cite{galton1884measurement, allport1936trait, goldberg1990big} (the idea that stable individual differences leave measurable traces in everyday language) APPT has been studied for decades and has benefited from major advances in representation learning. Yet it remains challenging in practice, because trait-related cues in language are subtle, unevenly expressed, and strongly shaped by context. Over time, approaches have ranged from dictionary-based feature extraction (e.g., Linguistic Inquiry and Word Count) and traditional supervised classifiers (e.g., Logistic Regression, Support Vector Machines, Random Forests), to neural encoders such as transformers (e.g., BERT) and, more recently, large language models (LLMs). Despite this progression in model complexity, it is often unclear whether apparent improvements reflect more faithful personality measurement or dataset- and protocol-specific effects, especially when evaluation reports only overall performance, or focuses on the presence of a trait, without showing whether the model can distinguish between the presence and absence of that trait. More broadly, the limited evidence from evaluations across different data sources suggests that models trained in one setting may not generalize reliably to another, as their predictions remain sensitive to the type of text, the users represented, and the statistical patterns of the dataset. This casts doubt on strongly nomothetic formulations of the task when they are justified only through aggregate metrics, and suggests that models may need to be evaluated (and in some cases adapted) with greater sensitivity to context, population, and the distribution of personality-related evidence. This concern is reinforced by a recent study: APPT with zero-shot LLMs showed that performance varies across models and datasets, and that aggregate metrics can mask substantial class-specific biases\cite{di2025mind}. These findings highlight the need to evaluate personality prediction beyond aggregate performance, including when using LLMs.

Evaluating whether models can reproduce an individual's personality requires a standardized representation of personality traits. The Big Five is one of the most widely used frameworks for organizing broad individual differences~\cite{john2008paradigm}. It represents personality along five continuous dimensions: Openness, Conscientiousness, Extraversion, Agreeableness, and Neuroticism. The framework has its roots in the lexical tradition, which identified and organized personality-descriptive terms in natural language~\cite{allport1936trait}. Subsequent factor-analytic research found a recurring five-factor structure across broad sets of such terms~\cite{goldberg1990big}. For these reasons, we adopt the Big Five as the personality framework for this study.

Against this background, we recast APPT through a more demanding question: \emph{can an LLM act as a digital twin of an individual’s personality?} Under this view, a model should not merely output plausible trait values on average; it should generate questionnaire-style responses consistent with a specific person, including individuals in the tails of the trait distribution. If sufficiently reliable, personality digital twins could support scalable analyses of how behavior varies across personality profiles. They could enable controlled comparisons among different profiles without requiring the recruitment of large groups of real users for every study, which is often costly and time-consuming. This would make them particularly useful for exploratory analyses and large-scale simulations, provided that they reproduce profile-specific behavior rather than generic population tendencies.

To operationalize this idea, we adopt a Questionnaire-Conditioned formulation of APPT (QC-APPT). To the best of our knowledge, this is the first work to place questionnaire answering at the center of evaluating LLMs as personality digital twins, asking whether a model can answer as a specific individual rather than fall back on population-level tendencies. We examine this ability across complementary forms of person-specific information: complete personal writing, progressively expanding text context, and explicit personality profiles. Concretely, instead of predicting traits directly from text, the model answers individual Big Five questionnaire questions based on a user’s writing, and these answers are then aggregated into continuous trait scores. For each user, we concatenate Online Social Network (OSN) posts\footnote{We use the Facebook-based MyPersonality dataset for the main analysis.} into a single long document and condition the model on the questionnaire items. Given the user document and an item, the model predicts the user’s response on a five-point Likert scale.

Crucially, we evaluate whether the model’s answers reflect person-specific information rather than general expectations about how people usually respond. We therefore compare its behavior with both a model that generates answers by sampling  a value from the distribution of scores on the Likert scale obtained from ~\cite{srivastava2003development} and the model’s own default responses in the absence of personal text. We also test whether performance remains reliable for individuals with very high or very low trait scores, where average-looking results may hide weak personalization. To assess the role of context, we measure how performance changes as progressively larger portions of user text are provided, asking whether trait cues remain stable or become diluted as context grows. Finally, we include an explicit-profile setting in which the LLM receives no user text and is provided only with the individual's Big Five profile, not as a standard personality prediction scenario, but as a stress test of digital twinning under direct psychological guidance. By providing the target Big Five profile without user text, we assess to what extent the LLM can interpret the Big Five traits and express them consistently through questionnaire responses.

To answer our overarching question, we investigate three research questions: \textbf{RQ1:} \emph{Can LLMs use user-generated text to answer Big Five questionnaire items in a way that reproduces the target individual's ground-truth personality profile more faithfully than population-level and model-internal priors?} \textbf{RQ2:} \emph{When given a target Big Five profile directly, can LLMs produce questionnaire responses that consistently reproduce that profile across low, intermediate, and high trait levels?} \textbf{RQ3:} \emph{How does the amount of available user-generated text affect the accuracy and stability of an LLM's questionnaire-based personality reproduction?}

\noindent Our main contributions are:
\begin{itemize}
    \item We propose a questionnaire-conditioned framework for evaluating LLMs as personality \emph{digital twins}, in which models answer Big Five questionnaire and their responses are aggregated into continuous trait scores.

    \item We introduce a distribution-aware evaluation that examines performance not only on average, but also for individuals with very low or very high trait scores, showing when average error hides weak personalization.

    \item We distinguish genuine personalization from general response tendencies using two complementary baselines: population-level expectations based on the Srivastava trait distribution~\cite{srivastava2003development} and a no-text LLM setting that captures the model’s default answers without personal evidence.

    \item We introduce an explicit-profile stress test, where the LLM is given only a target psychological profile, to assess whether it can translate direct personality information into consistent questionnaire answers.

    \item We analyze how the model’s ability to mirror a user’s questionnaire behavior changes as more personal text becomes available and as relevant information appears in different parts of the input.
\end{itemize}

\noindent The remainder of the paper is organized as follows. Section~\ref{sec:related} reviews related work on APPT and persona/digital-twin modeling. Section~\ref{sec:method} describes the questionnaire-conditioned formulation, baselines, and evaluation metrics. Section~\ref{sec:experiments} presents datasets, models, and experimental settings, including the increasing-context protocol. Section~\ref{sec:results} reports results and analyses, and Section~\ref{sec:conclusion} concludes with limitations and implications for deploying LLM-based personality digital twins.

\section{Related Work}
\label{sec:related}

Research on computational personality modeling is grounded in a much older psychometric tradition. Within personality psychology, the lexical program of Allport and Odbert catalogued trait-descriptive terms in natural language\cite{allport1936trait}, while Goldberg later articulated the Big Five factor structure as a broad account of personality variation \cite{goldberg1990big}. In turn, questionnaire design studies such as the public-domain IPIP inventory \cite{Goldberg1999_IPIP,Johnson2014_IPIP120}, the BFI-2 family \cite{SotoJohn2017_BFI2,SotoJohn2017_BFI2Short}, and the TIPI \cite{Gosling2003_TIPI} operationalized that construct into item-based instruments with different trade-offs between breadth, facet coverage, and brevity.
%This measurement perspective matters for the present paper because the target is not simply a trait label, but questionnaire-consistent behavior at the item level.

The NLP literature on APPT established that linguistic production contains trait-related signal, but also showed that the task is subtle and highly dependent on data and evaluation design. Early influential work by Mairesse et al.\ reported experimental results for recognition of all Big Five traits in both conversation and text, using self- and observer-rated personality\footnote{More on the Brunswik’s Lens Model and the Kernel of Truth can be found in \cite{brunswik2023perception,funder1995accuracy}} and comparing multiple modeling settings\cite{mairesse2007using}. In parallel, large-scale social media studies demonstrated that digital traces can support the prediction of personal attributes, including personality-related variables, from online behavior \cite{kosinski2013private}. Subsequent work on social media language further examined how personality is expressed in naturally occurring online writing and how much text may be needed before prediction becomes more stable \cite{qiu2012you,arnoux201725}. At the same time, the literature has repeatedly noted that results depend strongly on platform, labeling strategy, and task formulation.

A more specific strand of work has brought psychometric structure into personality prediction by using questionnaire items as an intermediate scaffold for inference. Yang et al.\ introduced PQ-Net, a two-stream architecture for personality detection in which a questionnaire stream guides a pretrained text encoder toward post content relevant to the target traits \cite{yang2021learning}. More recently, PsyCoT framed questionnaire items as a structured chain-of-thought process, prompting an LLM to rate items in multiple turns based on the user’s texts before deriving a final personality preference \cite{yang2023psycot}. These studies are especially close to the present work in that they make questionnaire structure operational within text-based personality modeling. At the same time, they remain centered on personality detection. In particular, PsyCoT is studied on the Essays benchmark in a binary formulation, where the final target is a high/low personality label and performance is reported with classification metrics such as accuracy and macro-F1. More broadly, this evaluation setting remains close to the coarse-grained classification perspective documented in earlier APPT literature, where performance is typically summarized through aggregate metrics rather than through distribution-aware analysis. Item-level 1--5 ratings are used as intermediate evidence and as a secondary consistency check with questionnaire-defined scoring rules, rather than as the primary psychometric output under evaluation. Moreover, PsyCoT does not analyze whether questionnaire-based predictions remain distinguishable from prior-based baselines,\footnote{We use \emph{prior} to refer to general response tendencies that may influence personality predictions independently of the target individual, including population-level regularities in how people usually score and the model's own default answering behavior when no personal information is available.} nor does it examine text-length effects, performance across the ground-truth trait distribution, or questionnaire-answering behavior under explicit personality-profile conditioning.

A related but distinct body of work examines personality in language generation and in LLMs themselves. On the generation side, PERSONAGE introduced a parameterizable dialogue generator that produces utterances varying along the extraversion dimension, as validated by human judgments \cite{mairesse2007personage}. In recent LLM work, Caron and Srivastava studied whether generated text exhibits consistent perceived Big Five traits and whether such traits can be manipulated through context \cite{caron2023manipulating}. Jiang et al.\ introduced PersonaLLM to investigate whether prompted LLM-based agents can express target Big Five traits across questionnaire completion and text generation settings \cite{jiang2024personallm}. Sorokovikova et al.\ further examined the simulation and stability of Big Five traits in several contemporary LLMs \cite{sorokovikova2024llms}. Collectively, this literature shows that personality is now used both as an analysis lens and as a controllable property in generative systems. Yet the central question in those works is whether models can display or simulate personality-like patterns, not whether they can reproduce the questionnaire behavior of a given person from that person’s evidence under prior-aware and distribution-aware evaluation.

The present paper is positioned at the intersection of these traditions. Like APPT, it starts from user-generated text; like questionnaire-guided personality prediction, it incorporates psychometric items into the modeling process; and like recent LLM personality studies, it asks whether a model can enact a coherent profile. Its distinguishing feature is that questionnaire response generation is treated as the evaluation target itself, rather than as an intermediate device for improving trait prediction. Instead of asking whether a model predicts plausible trait scores on average, the paper asks whether an LLM can function as a personality digital twin by producing item-level questionnaire responses for a specific individual in a way that remains distinguishable from prior-based baselines and informative across the ground-truth score distribution. To support that question, the paper introduces a questionnaire-conditioned formulation, compares against both a population prior and a no-text model prior, evaluates performance across the ground-truth distribution rather than only in aggregate, and studies what happens when the model is conditioned either on user text or on an explicit psychological profile. Additionally, the study investigates the relationship between text length and predictive performance across all experimental settings. In this sense, the gap addressed here is not the absence of personality modeling in NLP or LLMs, but the lack of a distribution-aware, questionnaire-level evaluation of whether such models genuinely go beyond priors when asked to reproduce individual personality behavior.

\section{Methodology}
\label{sec:method}

Psychological constructs are traditionally assessed through standardized questionnaires, in which individuals respond to structured questions and their answers are aggregated into trait scores. We adopt this measurement principle in a computational setting: given user-generated text, we employ LLMs to answer Big Five questionnaire questions as a target user, derive trait scores from these predicted answers using the corresponding scoring keys, and evaluate whether these answers reflect the target user rather than general population tendencies or the model’s own default behavior, particularly for users with very low or very high trait scores. This questionnaire-based formulation is motivated by the assumption that personality inference may be more tractable when LLMs reason through short, focused questionnaire questions than when they must extract subtle trait cues directly from noisy natural text. It also aligns the task more closely with questionnaire-based personality assessment and allows us to evaluate the model’s answers question by question, rather than only through final trait scores.
As a secondary experiment, we also consider an explicit-profile setting with a distinct purpose: the model receives no user text, but only the target personality profile. This allows us to test whether direct psychological guidance can be translated into consistent questionnaire answers.

\subsection{How we set up the digital twin}

The leveraged ground truth contains two types of information: (i) user-authored posts and (ii) the corresponding Big Five personality profile. Based on these two modalities, we instantiate two experimental configurations with distinct scopes.

In the first configuration, the LLM receives the user’s \textbf{text as input} and answers to questionnaire items accordingly. This form of digital twin can be considered \textit{implicit}, as personality characteristics are inferred from long, noisy, and context-dependent linguistic signals. The model reconstructs the individual semantically, relying on distributed behavioral cues embedded in natural language.

In the second configuration, the LLM receives the structured \textbf{Big Five profile directly as input}. This version can be considered \textit{explicit}, as the model is conditioned on ground-truth psychometric scores and is prompted to generate responses consistent with those traits. Although this setting is not intended as a practical personality prediction scenario, it serves as a benchmark for testing whether LLMs can understand the behavioral implications of an explicit personality profile and express them consistently when completing the questionnaire.

While the text-based configuration evaluates the model’s ability to extract and operationalize personality cues from naturalistic data, the profile-based configuration allows us to assess whether the LLM can express an explicit psychometric profile through consistent questionnaire responses without relying on noisy textual evidence.

\subsection{How we ask the digital twin to answer the questionnaire}

LLMs are known to use long contexts imperfectly, with performance depending in part on where relevant information is placed within the prompt \cite{liu2024lost}. While large proprietary models can generally accommodate longer inputs, extended prompts may still be challenging, especially for smaller open-source models. For this reason, we adopt two different questionnaire administration strategies.

In the first configuration, the \textbf{entire questionnaire is provided within a single prompt (one-batch)}. This approach is computationally efficient because it requires one model call per user-questionnaire pair, although the length and cost of each call increase with the amount of user text and the number of questionnaire questions. However, it presents at least two potential risks: (i) ignored part of the prompt in models with limited context windows, and (ii) reduced attention to individual items, potentially affecting response quality.

In the second configuration, \textbf{each questionnaire item is submitted in a separate prompt (per-item)}. This strategy increases computational cost, as the number of requests scales with the number of texts, questionnaires, and items. Despite this overhead, it offers two advantages: it minimizes the risk of input shortening due to length constraints and allows the model to focus more explicitly on each individual item.

By comparing these two administration strategies, we assess whether prompt granularity affects the stability and reliability of the digital twin’s responses. This distinction is particularly relevant when evaluating heterogeneous models with different context capacities. The comparison allows us to disentangle performance differences due to personality modeling from those caused by prompt-length sensitivity.

\subsection{Different types of questionnaires}

We consider several widely used Big Five questionnaires that are publicly available and frequently adopted in the personality literature. These instruments differ in length and level of psychometric resolution, ranging from long-form inventories to ultra-short scales. Among them, we focus the main experiments on the \textbf{Big Five Inventory-2-60} (BFI-2--60)\cite{SotoJohn2017_BFI2}. This questionnaire offers a balanced compromise between long-form inventories and ultra-short scales: it is detailed enough to preserve meaningful psychometric structure, while remaining compact enough for a study that requires repeated model calls across users, datasets, and experimental conditions. This choice is also supported by preliminary experiments with shorter variants, where reducing questionnaire length tended to increase prediction error.  For clarity of presentation, the main analysis therefore reports results on BFI-2-60.

The number in the questionnaire name refers to its number of items, and the number of items per trait can be obtained by dividing that number by 5 (i.e., the number of Big Five traits). The \textbf{BFI-2-60}, the revised version of the BFI, comprises 60 items, with 12 items measuring each trait. It follows a hierarchical structure comprising fifteen facets (i.e., narrower personality dimensions) organized under the five broad Big Five domains (i.e., traits), designed to capture personality in enough detail without requiring an excessively long questionnaire. Its abbreviated forms, the \textbf{BFI-2-S (30 questions)} and \textbf{BFI-2-XS (15 questions)}, preserve the core domain structure while progressively reducing facet-level resolution and respondent burden \cite{SotoJohn2017_BFI2Short}. These versions offer a structured compromise between psychometric precision and efficiency\footnote{All BFI-2 questionnaire variants and scoring keys can be found at \url{https://www.ocf.berkeley.edu/~johnlab/measures.html} (The Big Five Inventory 2).}.

At the longer end of the spectrum, the \textbf{International Personality Item Pool-300} (IPIP-300) is grounded in Goldberg’s public-domain broad-bandwidth inventory, designed to measure lower-level facets of the Big Five through extensive item coverage \cite{Goldberg1999_IPIP}. The \textbf{IPIP-NEO-120} further operationalizes this framework by providing a 120-item instrument that captures thirty facets (six per trait), offering detailed and psychometrically robust domain- and facet-level assessment while reducing administration length \cite{Johnson2014_IPIP120}. These instruments emphasize internal consistency and fine-grained trait representation\footnote{IPIP-300 and IPIP-120 questionniare items and scoring keys can be found at \url{https://osf.io/ycvdk/files/osfstorage} (IPIP-NEO-ItemKey.xls).}.

At the shortest end of the spectrum, the \textbf{TIPI-10} \cite{Gosling2003_TIPI} provides an ultra-brief assessment with two questions per trait. It is explicitly designed for contexts where time constraints preclude longer inventories, accepting reduced internal redundancy in exchange for rapid administration\footnote{The TIPI-10 questionnaire and its scoring keys can be found at \url{https://gosling.psy.utexas.edu/wp-content/uploads/2014/09/tipi.pdf}}.

For experimental consistency, all questionnaire items were standardized into a uniform first-person ("I") declarative format while preserving their original semantic content. Questionnaires originally distributed as PDF documents (e.g., BFI-2 and TIPI-10) were manually converted into structured CSV format to enable automated prompting and scoring.

This range of instruments allows us to examine how the digital twin behaves under varying levels of psychometric density and measurement precision, from redundancy-rich inventories to highly compressed trait representations.

\section{Experiment settings}
\label{sec:experiments}

\begin{table*}[!t]
\centering

\setlength{\tabcolsep}{5pt}
\resizebox{\linewidth}{!}{
\begin{tabular}{|c|cc|cc|ccccc|}%ccccc|
\toprule
\textbf{Dataset} & \textbf{\#Texts} & \textbf{\#Users} 
& \textbf{Char Length} & \textbf{Word Length} 
& \textbf{OPN} & \textbf{CON} & \textbf{EXT} 
& \textbf{AGR} & \textbf{NEU} 
%& \textbf{OPN (1/0)} & \textbf{CON (1/0)} 
%& \textbf{EXT (1/0)} & \textbf{AGR (1/0)} 
%& \textbf{NEU (1/0)} \\
\\
\midrule
mypersonality & 125 & 125 
& 4627.0 $\pm$ 4266.11 
& 867.0 $\pm$ 787.59 
& 4.25 $\pm$ 0.62 
& 3.45 $\pm$ 0.76 
& 3.40 $\pm$ 0.85 
& 3.75 $\pm$ 0.65 
& 2.60 $\pm$ 0.75 \\
%& 93 / 32 
%& 62 / 63 
%& 52 / 73 
%& 71 / 54 
%& 47 / 78 \\
\bottomrule
\end{tabular}
}
\vspace{0.1cm}
\caption{Summary statistics for the MyPersonality dataset after merging posts per author id. Continuous traits are reported as Median $\pm$ SD.}
\label{df_summary}
\end{table*}

The MyPersonality dataset is one of the most widely used resources for online writing.\footnote{MyPersonality dataset available at \url{https://github.com/sridharsrinivasan7/MyPersonality_NLP/blob/master/Dataset/mypersonality_final.csv}} Developed by Stillwell and Kosinski~\cite{kosinski2013private}, the original Facebook app collected user responses to Big Five and MBTI questionnaires. Many users voluntarily shared both their personality scores and behavioral data for research purposes. The dataset was extensively used until around 2018, when it was withdrawn due to privacy concerns around the time of the Cambridge Analytica scandal. A subset of 10,000 entries from 250 distinct users remains publicly available on platforms such as GitHub\footnote{Mypersonality data comprises the author ID, the very short posts along with continuous and binary labels of BIG5 traits, and some network topology metrics such as network size, betweennes and density}.

To ensure sufficient linguistic material per user, we aggregate posts by author ID and retain only users with at least 25 posts. Prior work shows that trait variability decreases sharply as sample size increases from approximately 20 to 100 tweets, before stabilizing more gradually around 200 tweets (see Figure 3 in \cite{haber2015stability}). While highly stable percentile estimates may require larger samples (e.g., ~200 tweets or 5,000 words), the most substantial reduction in variance occurs in the early range. In this context, a threshold of 25 posts provides a meaningful reduction in sampling variability while preserving adequate dataset coverage, consistent with prior Twitter-based personality studies \cite{arnoux201725}. The summary of the resulting dataset is shown in Table~\ref{df_summary}.

\subsection{LLMs considered}

We evaluate the following LLMs:
\begin{itemize}
    \item \texttt{GPT-4} ($\sim$1T parameters)\footnote{This is an estimate\cite{Strickland2024WhatIsGenerativeAI}, given that OpenAI did not release the number of parameters.}
    \item \texttt{gemma:7b-instruct} (7B parameters)  
    \item \texttt{llama3.1:8b} (8B parameters)  
    \item \texttt{mistral:instruct} (7B parameters)  
\end{itemize}
Open-source LLMs are accessed via Ollama\footnote{Ollama is an open-source framework for downloading and running LLMs locally. See the GitHub repository at: \url{https://github.com/ollama/ollama}}.
Temperature is set to 0 to make the generations as deterministic as possible. As reported in the Results section, repeated runs produced nearly identical outcomes, with little to no variation. Any residual variation may depend on other generation settings, such as top\_p and top\_k, which were left at their default values.
Finally, the prompt templates used for all variants are reported in the Appendix. Each prompt is composed of a system message, which defines the role, allowed response scale, and output rules, and a user message, which provides either the user text and questionnaire content or only the questionnaire content. The variants differ only in the information provided and in the required output format: text-conditioned prompts ask the model to answer as the author of the text, while no-text prompts capture the model's default answering behavior; one-batch prompts require answers to the full questionnaire in a single JSON-like output, while per-item prompts require one allowed response for one question at a time.

\subsection{Baselines}

In order to assess the performance of LLMs as digital twins, we leverage different baselines whose description is provided in the following.

\subsubsection{Distribution based}
To contextualize the distribution of personality traits in our dataset, we adopt as a normative reference the large-scale Internet sample reported by Srivastava et al.\cite{srivastava2003development}, hereafter referred to as Srivastava. Their data were collected through the Berkeley Personality Lab website in the late 1990s and early 2000s using the 44-item Big Five Inventory (BFI). The sample comprised over 130,000 self-selected participants, predominantly from the United States but spanning a broad range of ages and backgrounds. Trait scores were computed as the mean of item responses on a 1–5 Likert scale. The reported distributions were approximately normal, with means slightly above the midpoint and standard deviations typically ranging between 0.6 and 0.8. Owing to its scale and methodological rigor, this dataset provides one of the earliest and most widely cited normative benchmarks for Internet-based Big Five assessment.

We employ two distinct normative baselines derived from this reference: \textbf{SrivastavaDistr} and \textbf{SrivastavaMean}. \textbf{SrivastavaDistr} is the main baseline and is based on the simulated trait distribution: for each instance, it samples a value from the reconstructed Srivastava distribution. This makes it a stochastic baseline that reflects the reference population. By contrast, \textbf{SrivastavaMean} is deterministic: it always assigns the reported population mean for each trait and therefore captures only central tendency, without representing the full population spread. In addition, we introduce a deterministic midpoint baseline that assigns the midpoint value (i.e., 3; \textbf{Center}) to all instances, enabling comparison against a psychologically neutral, scale-centered reference that is independent of population priors. The SrivastavaDistr baseline therefore allows us to test whether providing user text improves prediction over simply sampling from a general population distribution.

\subsubsection{LLM null model}
Beyond population-level priors, we also investigate LLM-internal priors, i.e., the trait distributions implicitly learned by the model during training. To estimate these, we reproduce each experimental configuration without providing user text in the prompt. In this setting, the LLM responds to the questionnaire items relying solely on its internal statistical tendencies, thereby revealing its baseline trait imitation in the absence of input-specific signals.

This procedure allows us to assess the extent to which textual information influences LLM outputs. In particular, we can determine whether the model meaningfully conditions its predictions on the provided user text, or instead defaults to its internal priors. Furthermore, this setup enables a comparison between early Internet population norms (Srivastava) and the implicit priors embedded in contemporary LLM training data.

Taken together, the Srivastava distribution, the no-text LLM baseline, and the central midpoint baseline provide a triangulated evaluation framework for assessing LLM performance when text is supplied.

\subsection{Evaluation metrics}

The most common metric for evaluating prediction with continuous values is the Mean Absolute Error (\textbf{MAE}), which directly quantifies the average deviation between predicted and ground-truth scores. While MAE provides an intuitive measure of error, as an aggregate statistic it may mask systematic patterns in the distribution of errors. In particular, models that exhibit regression-to-the-mean behavior (a well-documented phenomenon in continuous prediction settings) may achieve competitive overall MAE while underestimating extreme values.

To avoid this bias, we complement overall MAE with a distribution-aware evaluation. Specifically, we compute MAE separately across three percentile ranges of the ground-truth distribution: 0–25\% (lower tail), 25–75\% (central region), and 75–100\% (upper tail). This stratified analysis allows us to assess whether the model meaningfully captures extreme cases or primarily predicts values close to the central tendency. In this way, performance in the tails can be directly compared against the no-text LLM baseline (model prior) and the Srivastava reference distribution (population prior), yielding a more granular assessment of predictive behavior.

Although correlation coefficients are frequently reported in continuous prediction tasks, they are less informative in the present setting. Given the restricted 1–5 Likert scale, the concentration of scores around the midpoint, and the relatively limited variance in predictions, correlation may appear attenuated when the absolute errors are small. For this reason, we emphasize error-based metrics, which more directly reflect predictive accuracy on the original measurement scale.

Finally, we do not report RMSE. While RMSE penalizes larger deviations more strongly than MAE, our percentile-conditioned MAE already provides explicit insight into performance on extreme values, thereby addressing the primary concern that RMSE is typically intended to capture.

\subsection{Statistical Reliability}
To assess variability across runs, multiple runs are performed for each experiment. In the main analysis (i.e., full-text settings), five runs are conducted per experiment. For experiments with incremental context window, three runs are used instead, due to the larger number of configurations introduced by the sliding window setup. All reported confidence intervals are computed at the 95\% level.

The analysis focuses on the BFI-2-60, as it provides a balanced trade-off between the higher granularity of the IPIP-300 and the compactness of the TIPI-10. The PerItem setup is particularly demanding because it requires a separate model call for every questionnaire item; therefore, the text-conditioned PerItem configuration with GPT-4 was evaluated only with BFI-2 questionnaires, while repeated runs of the most demanding IPIP-300 configurations were avoided. Repeated runs were instead retained for BFI-2 questionnaires to ensure consistency of comparison. Experiments involving the remaining questionnaire variants (IPIP-300, IPIP-120, BFI-2-30, BFI-2-15, and TIPI-10) were conducted but are omitted from the main presentation to preserve clarity of exposition. All plots produced for these configurations, including aggregate-MAE, ground-truth-region, explicit profile variant, and failure analysis, are reported in the accompanying technical report~\cite{TECHREP}. These supplementary results broadly confirm the conclusions obtained with BFI-2-60 and indicate that error tends to become larger and more variable for the shortest questionnaires, although this pattern is not uniform across all models, traits, and configurations. Failures correspond to invalid outputs from open-source LLMs and safety-filtered responses from GPT-4; they are limited in many settings but concentrated in specific model--variant--questionnaire combinations. Finally, in the sliding context window variant, Gemma produces invalid outputs when handling contexts longer than approximately 3{,}500 words.

\section{Results}
\label{sec:results}

This section evaluates whether LLMs behave as personality digital twins, namely, whether they produce questionnaire responses that reflect the target individual rather than generic expectations about how people usually score. The results are organized around four empirical questions. We first analyze text-based prediction in aggregate, asking whether user text improves over population-level and model-internal priors. We then examine whether those aggregate results remain valid across the full trait distribution, especially for people with relatively low or high trait values. Next, we test a different setting in terms of input, in which the model is given the ground-truth Big Five profile directly, in order to assess whether explicit psychometric information is sufficient to induce digital-twin-like behavior. Finally, we study whether increasing text length improves prediction.

Across all analyses, one conclusion remains stable: low average error is not, by itself, evidence of individualized behavior. In many configurations, models remain close to population-level or model-internal priors. The clearest positive results appear only in a limited number of model- and trait-specific cases, with GPT-4 standing out as the most convincing, but still bounded, exception.

\subsection{Text-based prediction: average performance is often prior-driven}

\begin{figure*}[!h]
    \centering
    \includegraphics[width=1\linewidth]{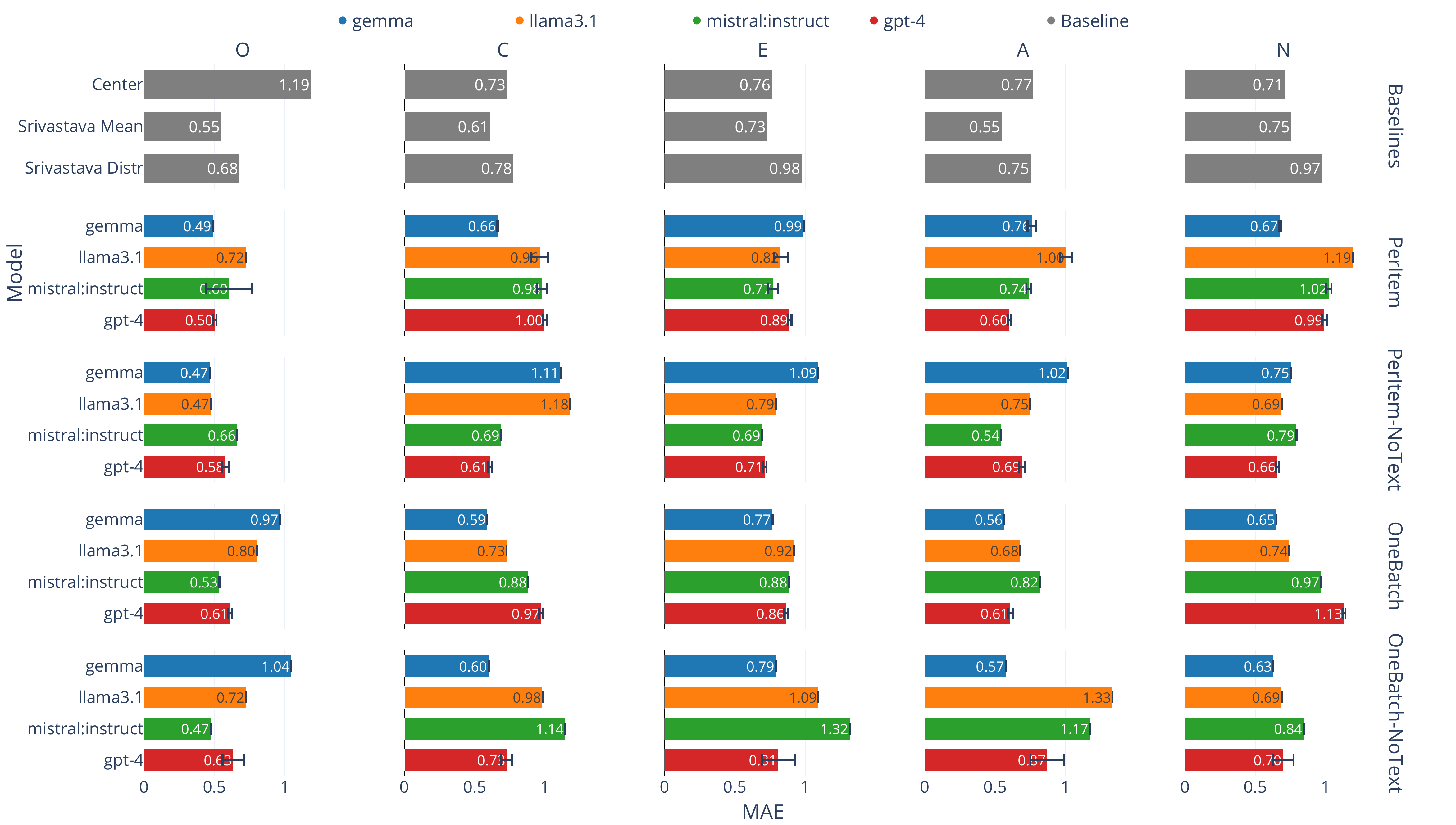}
    \caption{Mean absolute error per trait, model and variants --- text-based prediction, BFI-2-60 questionnaire only.}
    \label{fig:mae_hbars_mypers_bfi260}
\end{figure*}

We begin with Mean Absolute Error (MAE), which summarizes the average distance between predicted and ground-truth trait scores. Figure~\ref{fig:mae_hbars_mypers_bfi260} is organized by trait in columns and by prediction setting in rows. The top row reports the baselines: Center, SrivastavaMean, and SrivastavaDistr. The remaining rows report text-conditioned and no-text variants under the PerItem and OneBatch prompting strategies. Within each panel, bar color identifies the model, with GPT-4 shown in red and the open-source models in the other colors.

Figure~\ref{fig:mae_hbars_mypers_bfi260} shows that text-conditioned predictions are uneven across traits, models, and prompting formats, and that this unevenness must be interpreted against the baseline row. The SrivastavaDistr baseline is the main reference because it represents a sampled population-level expectation, whereas SrivastavaMean and Center provide deterministic diagnostic references: the former always predicts the population mean, and the latter always predicts the scale midpoint.
Against SrivastavaDistr, text-conditioned results frequently achieve a lower MAE, with margins that vary across configurations. Examples include Openness in the PerItem setting (second row), where Gemma and GPT-4 obtain 0.49 and 0.50 against 0.68 for SrivastavaDistr, and Neuroticism in OneBatch (fourth row), where Gemma obtains 0.65 against 0.97. Other cases also show an aggregate advantage over the main population-based baseline, but this advantage alone does not establish that the predictions exploit person-specific information.
The no-text variants illustrate this limitation. Low error is not limited to settings where user text is provided: for example, in PerItem-NoText (third row), Gemma and Llama obtain 0.47 in Openness against 0.68 for SrivastavaDistr, and Mistral obtains 0.54 in Agreeableness against 0.75. Similarly, in OneBatch-NoText (fifth row), Gemma obtains 0.57 in Agreeableness and 0.63 in Neuroticism, both below the corresponding SrivastavaDistr values. Additional cases show the same tendency. Thus, low aggregate error can arise even when no user-specific text is available.
The comparison between text-conditioned and no-text settings therefore suggests a limited and selective role for user text. The clearest text-conditioned gains appear in OneBatch, especially for Llama in Agreeableness, where error decreases from 1.33 without text to 0.68 with text, and for Mistral in Extraversion, from 1.32 to 0.88. By contrast, PerItem gains are generally smaller, as in GPT-4 for Openness (0.58 to 0.50) and Agreeableness (0.69 to 0.60). Other improvements exist, but the overall pattern is not systematic: several comparisons are nearly identical, and in some cases no-text predictions perform better.

Overall, at the aggregate level considered in Figure~\ref{fig:mae_hbars_mypers_bfi260}, user text appears to help in selected model--trait combinations, but not systematically. The clearest text-conditioned advantages are more pronounced in OneBatch than in PerItem. At the same time, low average error is not sufficient evidence of person-specific use of the text, because similar or lower errors can be obtained by no-text variants and by population-level baselines.
Figure~\ref{fig:mae_hbars_mypers_bfi260} therefore shows why aggregate MAE is not sufficient on its own: competitive average performance does not establish that the model is using person-specific information. Moreover, aggregate MAE cannot reveal whether this performance is maintained for individuals located in different regions of the trait distribution. Since the digital-twin question concerns individuals rather than averages, we therefore move to a distribution-aware analysis.

\begin{figure*}[!t]
    \centering
    \includegraphics[width=1\linewidth]{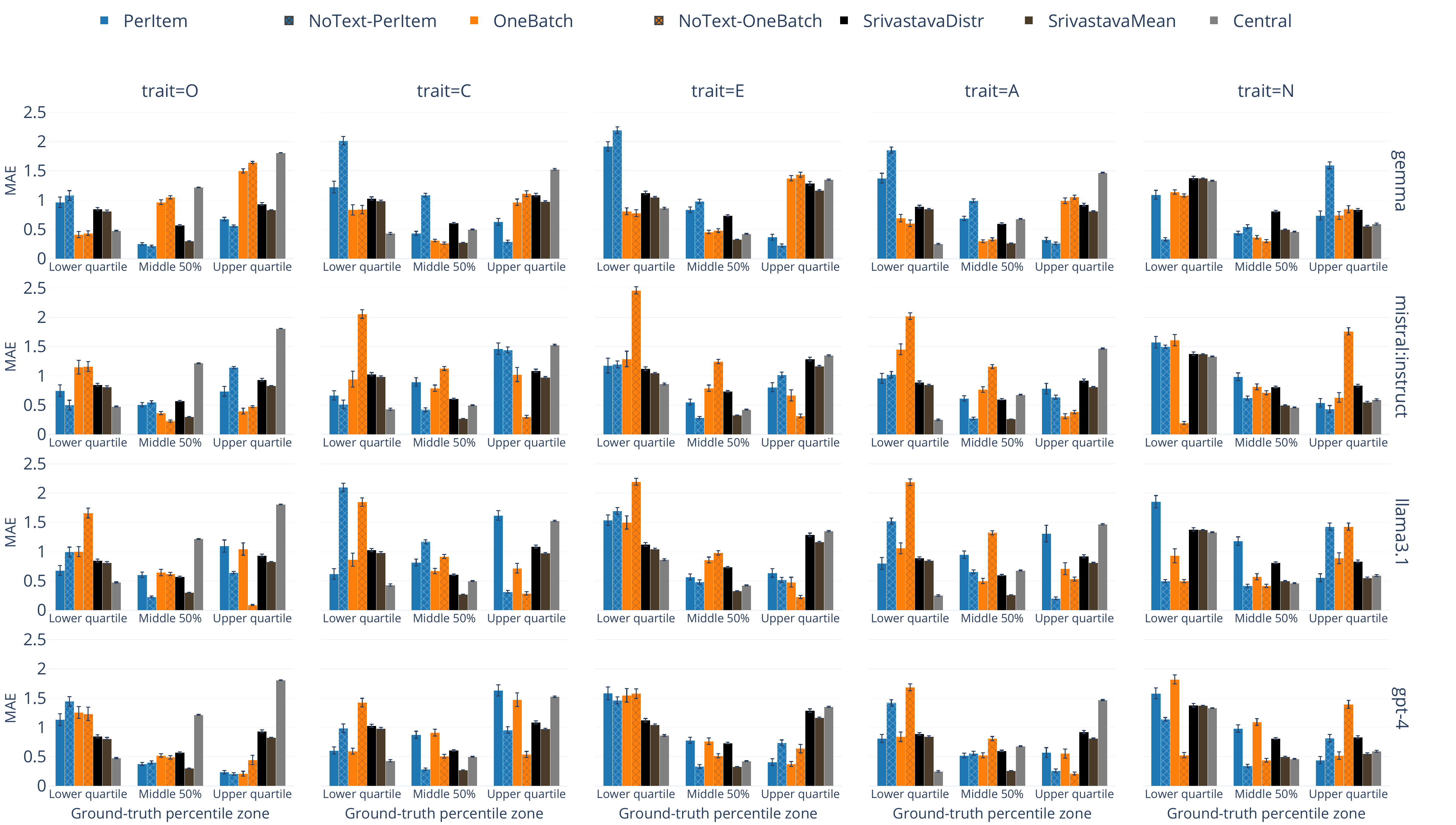}
    \caption{Mean absolute error per trait, model and variants conditioned on ground-truth percentile zones --- text-based predictions, BFI-2-60 questionnaire only.}
    \label{fig:mae_gt_bars_mypers_bfi260}
\end{figure*}

\begin{tcolorbox}[keyfindingbox, title=Key Finding 1:\\Low aggregate error does not establish personalization]
LLMs can appear reasonably accurate on average when predicting questionnaire scores from text, but competitive performance can also arise from population-level baselines or from the model's own default answering behavior. User text helps in selected model--trait combinations, especially in some OneBatch cases, but this benefit is not systematic. At the aggregate level, low error alone is therefore not sufficient evidence that the model is using person-specific information.
\end{tcolorbox}

\subsection{Text-based prediction across the distribution: the clearest successes are rare and uneven}

Figure~\ref{fig:mae_gt_bars_mypers_bfi260} opens up the aggregate result by reporting error separately for the lower quartile, the middle 50\%, and the upper quartile of the ground-truth distribution. Traits are arranged by columns, models by rows, and within each panel the bars compare the two text-based prompting strategies, their no-text counterparts, and the three baselines. This figure therefore asks a more demanding question than the previous one: can the model reproduce questionnaire behavior not only for average cases, but also for people with relatively low or high trait values?

The answer is mostly negative. The strongest text-based successes become rarer once the distribution is inspected directly. Many models remain reasonably competitive in the middle region but become less convincing at the lower and upper ends of the distribution. In plain terms, the models are better at producing plausible answers for more typical individuals than at reproducing how people with more distinctive trait levels would answer the questionnaire.

Among the evaluated models, GPT-4 provides the clearest positive examples, although these remain limited. One such case appears in the \emph{Agreeableness} column and the \emph{GPT-4} row, that is, the fourth column and fourth row of Figure~\ref{fig:mae_gt_bars_mypers_bfi260}. In that panel, the orange bar corresponding to the single-prompt administration remains competitive with the SrivastavaDistr baseline (black bar) and compares favorably with its no-text counterpart, especially in the lower quartile. Visually, this is one of the few panels in which textual conditioning appears to move the model away from a simple prior in a useful way across more than one region of the distribution. This makes GPT-4 the most convincing text-based case in the figure. At the same time, even GPT-4 does not show this behavior uniformly across all traits, so the correct interpretation is not that text-based digital twins are broadly working, but that the strongest model comes closer than the others in a limited number of cases.\\
Similar, though less striking, patterns also appear in a small number of additional panels. In Figure~\ref{fig:mae_gt_bars_mypers_bfi260}, this can be seen in the \emph{Conscientiousness} column for \emph{Llama} and \emph{Mistral}. Visually, the text-conditioned bars are not merely tracking their no-text counterparts or the population baseline. Rather, they separate from both references in meaningful parts of the distribution, showing that the user text is sometimes pushing the model toward a more individualized response pattern. At the same time, this advantage is not equally strong in all three regions, which is why these cases should be read as promising but still incomplete. They support the broader point that textual conditioning can help, but that clear and well-balanced gains across the full distribution remain relatively rare.\\
This figure also clarifies why average MAE can be misleading. A model may look acceptable in Figure~\ref{fig:mae_hbars_mypers_bfi260}, yet Figure~\ref{fig:mae_gt_bars_mypers_bfi260} can reveal a much weaker and less balanced form of personalization than the aggregate score suggests. The clearest example is GPT-4 under the PerItem setting for Openness: its aggregate MAE is 0.50, outperforming the SrivastavaDistr baseline of 0.68, but the distribution-aware analysis separates this result into an MAE of approximately 1.1 in the lower quartile, 0.4 in the middle region, and 0.2 in the upper quartile. The apparently strong aggregate result is therefore largely driven by very low error at the upper end of the distribution, while performance for individuals with low Openness remains substantially weaker and even exceeds the population-based baseline. Once the distribution is opened up, clearly successful text-based cases become both fewer and more trait-dependent. The most important implication of Figure~\ref{fig:mae_gt_bars_mypers_bfi260} is therefore not simply that performance worsens in difficult regions, but that personalization is uneven. Text may help in one part of the distribution without producing a similarly convincing effect elsewhere. For an individual-level task, this is a substantive limitation: the model is not yet reproducing questionnaire behavior in a robust and person-specific way across the full range of trait values.

\begin{figure*}[!t]
    \centering
    \includegraphics[width=1\linewidth]{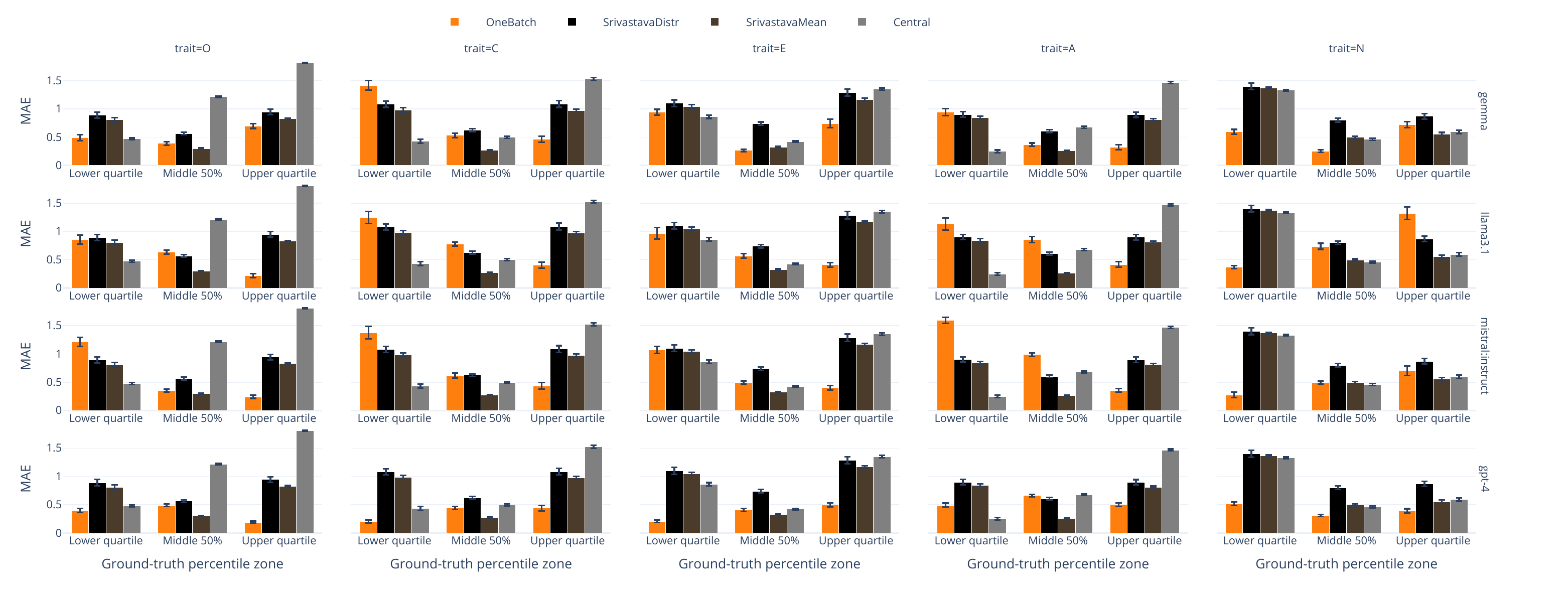}
    \caption{Mean absolute error per trait, model and variants conditioned on ground-truth percentile zones --- profile-based predictions, BFI-2-60 questionnaire only.}
    \label{fig:bar_gt_profile_preds}
\end{figure*}

\begin{tcolorbox}[keyfindingbox, title=Key Finding 2:\\LLMs struggle to replicate the behavior of extreme personalities]
LLMs are more convincing for people near the middle of the trait distribution than for those with unusually low or high trait values. In these more distinctive cases, predictions often drift back toward average, prior-like answers instead of preserving the person's profile. A small number of text-based exceptions do exist (most clearly GPT-4 on Agreeableness) but they remain too limited to change the overall pattern.
\end{tcolorbox}

\subsection{Profile-conditioned prediction: explicit personality information may help, but only GPT-4 uses it convincingly}

We now turn to a different and, in some sense, easier test. Instead of asking the model to infer personality from user text, we provide the ground-truth Big Five profile directly and ask the model to answer the questionnaire in a way that is coherent with that profile. Although this setup is not useful for personality prediction, it provides a stress-test to observe LLM capability to perform as a digital twin under explicit guidance. Figure~\ref{fig:bar_gt_profile_preds} uses the same layout as the previous figure: traits are arranged by columns, models by rows, and the three bars within each panel correspond to the lower quartile, middle 50\%, and upper quartile, alongside the same baselines. The difference is that the model is now conditioned on explicit profile information rather than on text.

Figure~\ref{fig:bar_gt_profile_preds} contains the clearest positive result in the profile-conditioned setting. This experiment is reported only under the OneBatch administration strategy because the input is a short, structured Big Five profile rather than long user-generated text; therefore, the context-length motivation for PerItem is absent, while OneBatch allows the complete profile to be expressed jointly across all questionnaire items. GPT-4 is able, in selected traits, to use the explicit profile in a genuinely informative way. The strongest cases are \emph{Extraversion} and \emph{Neuroticism}. Visually, these correspond to the third and fifth columns in the \emph{GPT-4} row. In both panels, the orange GPT-4 bars remain below the dark baseline bars not only in the middle region, but also in the lower and upper quartiles. This means that the model is not merely staying close to generic expectations around the center of the distribution, but is also retaining more of the target profile when questionnaire behavior becomes more distinctive.

These panels show that, under explicit profile conditioning, GPT-4 can translate psychometric information into questionnaire responses that are more faithful to the target individual. This is the clearest case in which the model moves beyond generic answering and behaves in a way that remains aligned with the provided profile across more than one part of the distribution.

At the same time, the same figure also shows why this result should be interpreted cautiously. First, it does not generalize to all models. In the rows corresponding to Gemma, Llama, and Mistral, the bars are often competitive in the center but remain much closer to the baselines in at least one of the outer regions. Visually, these panels do not show the same clear separation from the priors that GPT-4 shows in Extraversion and Neuroticism. Second, even for GPT-4 the effect is trait-dependent. In \emph{Openness}, \emph{Conscientiousness}, and \emph{Agreeableness} (the first, second, and fourth columns of the GPT-4 row) the profile-conditioned bars are still strong, but the advantage is not as uniformly clear as in Extraversion and Neuroticism.

This profile-based analysis clarifies where the main difficulty lies. The issue is not simply whether a model can generate questionnaire answers, but whether it can turn explicit personality information into a stable and faithful response pattern. Providing the correct profile helps, but for most models it is not enough. GPT-4 is the clearest exception, yet even there the success remains selective rather than general. Taken together, the profile-conditioned results sharpen the interpretation of the text-based ones. When the model must infer personality from text, strong evidence of individualized behavior is rare. When the profile is given explicitly, performance improves more clearly, but only GPT-4 uses that information in a consistently convincing way. This suggests that the problem is not only extracting personality from language, but also operationalizing personality information faithfully once it is available.

\begin{tcolorbox}[keyfindingbox, title={Key Finding 3:\\Knowing the personality profile is not enough, but GPT-4 approaches digital-twin behavior.}]
Most models do not reliably turn explicit Big Five information into questionnaire responses that remain faithful across the full distribution. GPT-4 is the clearest exception: when given the target profile, it most convincingly reproduces questionnaire behavior for Extraversion and Neuroticism, and more partially for the other traits. The result is promising, but still selective rather than general.
\end{tcolorbox}

\begin{figure*}[!h]
    \centering
    \includegraphics[width=1\linewidth]{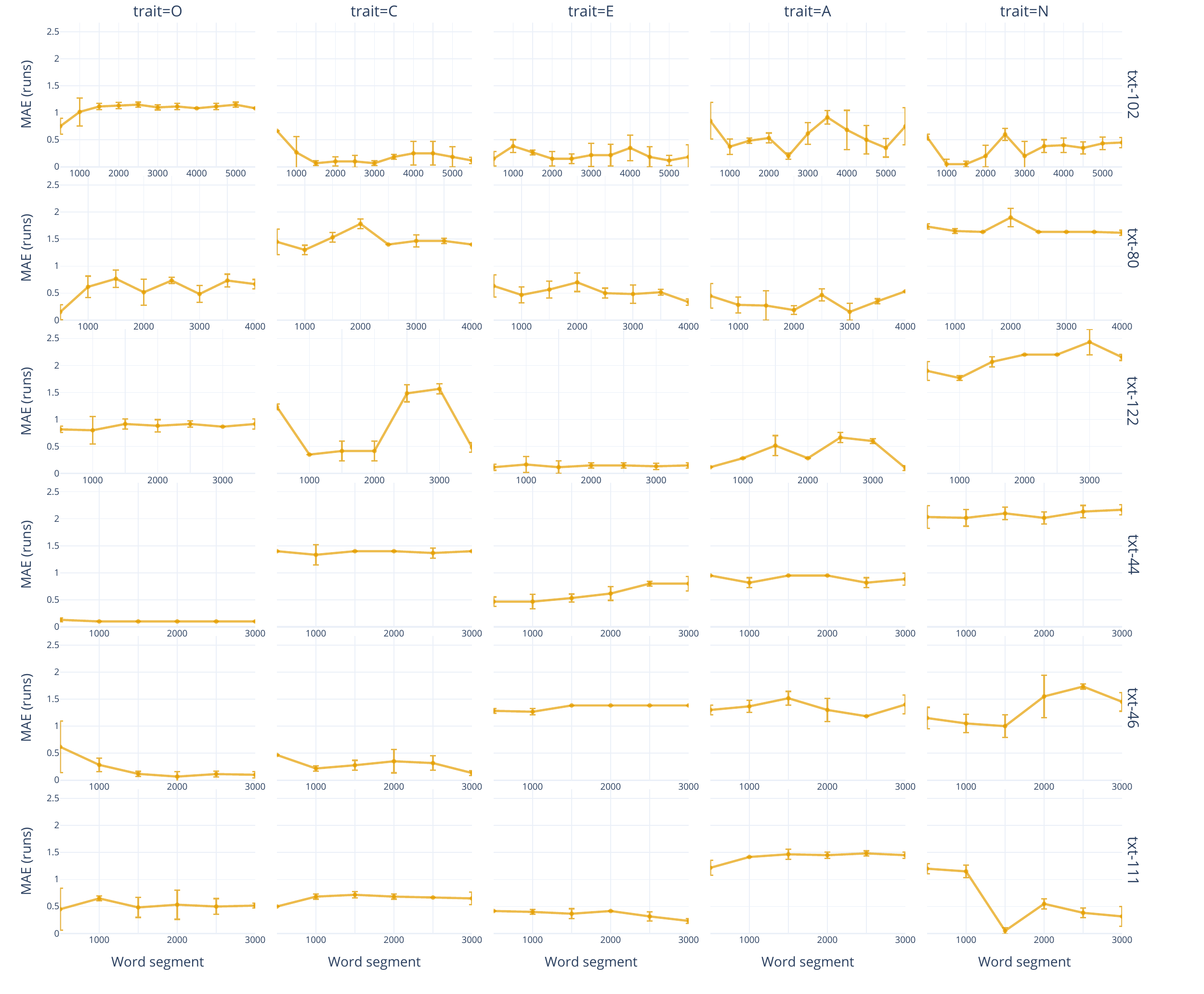}
    \caption{Absolute error per incremental word segment (GPT-4 only; six longest texts).}
    \label{fig:word_segment_ae_gpt_only_multi_txt}
\end{figure*}

\begin{figure*}[!h]
    \centering
    \includegraphics[width=1\linewidth]{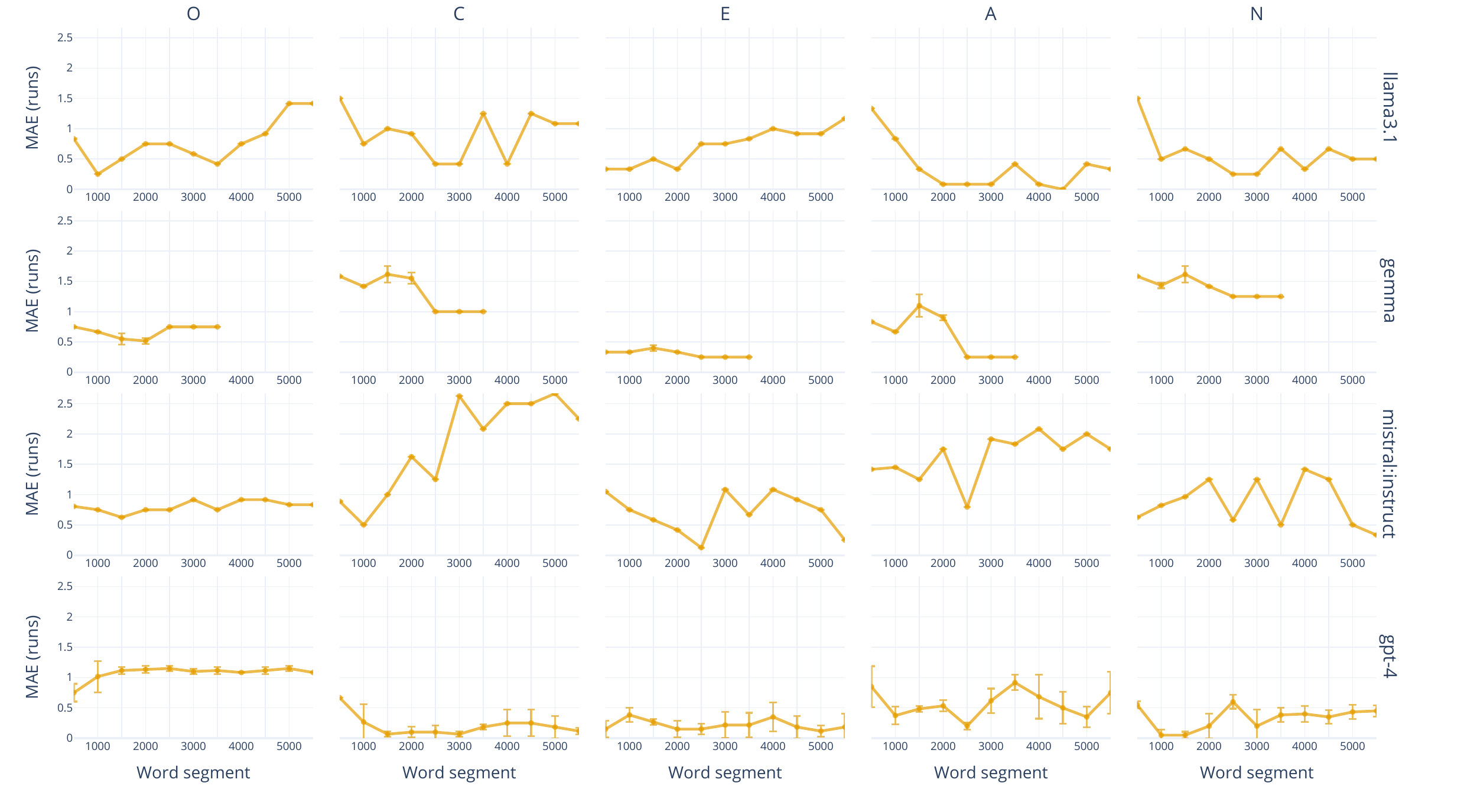}
    \caption{Absolute error per incremental word segment (all models; longest text).}
    \label{fig:word_segment_ae_all_models_longest_txt}
\end{figure*}

\subsection{Increasing text length does not reliably improve prediction}

A common assumption in personality prediction is that more text should provide more evidence and therefore lead to better estimates. Figures~\ref{fig:word_segment_ae_gpt_only_multi_txt} and~\ref{fig:word_segment_ae_all_models_longest_txt} show that this expectation is not supported consistently by the data.

Figure~\ref{fig:word_segment_ae_gpt_only_multi_txt} focuses on GPT-4 and reports six texts, one per row, ordered from longest to shortest. Traits are arranged by columns, and within each panel the line traces error as progressively larger text segments are added. The visual message of the figure is immediate: the lines do not move downward in a steady way as more text is provided. Some trajectories improve, others fluctuate, and others remain broadly flat. For example, in the \emph{Openness} column, one row shows a visible decrease as the text grows, while another shows instability rather than cumulative improvement. In the \emph{Neuroticism} column, some trajectories even worsen after additional context is added. The figure therefore does not support a simple ``more text is better'' interpretation.

Figure~\ref{fig:word_segment_ae_all_models_longest_txt} compares models on the same long text. Here, rows correspond to models and columns to traits. Again, the main result is visual and easy to read: there is no common downward trend across panels. What changes more clearly is the \emph{stability} of the trajectories. GPT-4, in the bottom row, shows smoother lines across segments, whereas the open-source models display sharper rises, drops, or oscillations in several traits. Thus, GPT-4 is not consistently better because more text always helps it, but because it appears to handle longer context more robustly. A practical limitation also emerges for Gemma. In the longest-context analyses, Gemma returns invalid outputs when texts exceed roughly 3{,}500 words, as the prompt is truncated and the model no longer reliably adheres to the required task.

The combined interpretation of the two figures is therefore clear. Additional context does not translate into a steady gain in personality prediction. Personality-relevant cues are not accumulated in a simple linear way, and more text can be helpful, neutral, or destabilizing depending on the segment and the model. GPT-4 again stands out, but mainly because it is more stable, not because it consistently improves with every additional block of text.

\begin{tcolorbox}[keyfindingbox, title=Key Finding 4:\\More text does not necessarily mean better personality prediction.]
Giving the model more text does not reliably make its questionnaire predictions more faithful to the person. Additional context can either help, make little difference, or even destabilize the prediction depending on the segment and the model. GPT-4 stands out mainly because its behavior remains smoother as context grows, whereas the open-source models are more erratic and, in Gemma's case, fail on very long inputs.
\end{tcolorbox}

\section{Conclusion}
\label{sec:conclusion}

This work asked whether an LLM can answer a personality questionnaire as a specific person would. We addressed this question by reframing personality prediction as a digital-twin problem, where success is not defined only by low average error, but by whether questionnaire responses remain person-specific and distinguishable from general response tendencies.

The results show that current LLMs do not yet behave as robust personality digital twins. In the text-based setting, apparent personalization is often limited: models frequently remain close to population-level tendencies or to their own default answering behavior. This becomes especially clear when performance is examined across the trait distribution. Models are more convincing near the center of the distribution, while individuals with very low or very high trait scores are more often pulled back toward average responses. A few text-based exceptions emerge, most notably GPT-4 on Agreeableness, but they remain selective rather than general.

Explicit personality profiles provide a stronger form of guidance, but they do not solve the problem for most models. GPT-4 is the clearest exception, especially for Extraversion and Neuroticism, where profile-conditioned responses more closely preserve the target profile across the distribution. Even in this setting, however, the effect remains trait-dependent. Increasing the amount of text also does not systematically improve prediction: additional context can help, make little difference, or destabilize the output depending on the model and text segment. Again, GPT-4 stands out as the most stable case, maintaining comparatively robust predictions as context increases, while the open-source models exhibit more erratic behavior.

Taken together, these findings suggest that current LLMs are better at reproducing broad personality tendencies than at faithfully mirroring individual questionnaire behavior. They can often generate average answers, but this does not necessarily imply digital-twin-like personalization. For this reason, evaluating personality digital twins requires more than aggregate metrics: average success can hide weak personalization, uneven behavior across the distribution, and a tendency to return toward what is typical. Future work should therefore focus on making person-specific behavior more robust for distinctive profiles, more stable under longer context, and more clearly separable from the general tendencies that models already bring with them.

\section{Limitations}
\label{sec:limitations}

This study has several limitations. First, the reported analysis focuses on MyPersonality. Experiments on Pandora were also conducted, but are not reported because the available user-level texts were considerably shorter and heavily preprocessed, limiting their suitability for the proposed evaluation. The findings should therefore be interpreted primarily within the MyPersonality setting.

Second, the reported experiments focus on BFI-2-60. Other questionnaire variants were also considered, but are omitted from the main presentation to preserve clarity. BFI-2-60 was selected because it offers the best trade-off between questionnaire length, psychometric granularity, and feasibility across the large number of experimental configurations.

Third, the model pool is selective. We evaluate three widely used open-source LLMs and GPT-4 as the proprietary reference model, balancing representativeness with tractability. Newer proprietary models may show stronger behavior, especially in long-context handling and in the explicit-profile setting.

\section*{Acknowledgment}
This work was partially funded by the European Union within the National Programme 'Research, Innovation and Competitiveness' 2021-2027, project SINTESI (Nr. MI\_0000962). F. Di Cursi's, C. Boldrini's, A. Passarella's, and M. Conti's work was also partly funded by the PNRR - M4C2 - Investimento 1.3, Partenariato Esteso PE00000013 - ``FAIR''

%\section*{References}
\bibliographystyle{IEEEtran}
\bibliography{bibliography}

\fi

\clearpage
\onecolumn

\appendix
\section*{Prompt variants}
\begin{lstlisting}[caption={Onebatch prediction with text}, label={onebatch_text}]
system_prompt=f"""You are a digital twin that fully adopts the perspective, preferences, and communication style of the person who wrote the provided text.
Your identity is defined ONLY by that text.
Given a list of questionnaire items, you need to answer to each item and return a json-like output (each key is the number of the item, each value is an answer).
Your ONLY allowed answers to each item are one of these exact strings:
- Disagree strongly
- Disagree a little
- Neutral
- Agree a little
- Agree strongly
- NA

Hard rules:
1) Output exactly one json output with exactly one of the six allowed strings above for each of the questionnaire items (NOT TRUE OR FALSE).
2) Do not output numbers, punctuation, explanations, qualifiers, or extra words.
3) If the provided text is too short/irrelevant/uninformative to answer the question reliably, output NA.
4) The output must be a json with the same length of the provided questionnaire, and one answer for each of the provided item, with item in the same order of the provided questionnaire.
"""

user_prompt=f"""TEXT:
{txt}

QUESTIONS (answer about yourself as the author and return a single json with all the answers for all questionnaire items):
{item_list}
"""      
\end{lstlisting}

\begin{lstlisting}[caption={Onebatch prediction without text}, label={onebatch_notext}]
system_prompt=f"""Given a list of questionnaire items, you need to answer to each item and return a json-like output (each key is the number of the item, each value is an answer).
Your ONLY allowed answers to each item are one of these exact strings:
- Disagree strongly
- Disagree a little
- Neutral
- Agree a little
- Agree strongly
- NA

Hard rules:
1) Output exactly one json output with exactly one of the six allowed strings above for each of the questionnaire items (NOT TRUE OR FALSE).
2) Do not output numbers, punctuation, explanations, qualifiers, or extra words.
3) If you don't know how to answer, output NA.
4) The output must be a json with the same length of the provided questionnaire, and one answer for each of the provided item, with item in the same order of the provided questionnaire.
"""

user_prompt=f"""QUESTIONS (answer about yourself and return a single json with all the answers for all questionnaire items):
{item_list}
"""
\end{lstlisting}

\clearpage
%%%%%%%%
\begin{lstlisting}[caption={Per item prediction with text}, label={peritem_text}]
system_prompt=f"""You are a digital twin that fully adopts the perspective, preferences, and communication style of the person who wrote the provided text.
Your identity is defined ONLY by that text.
Your ONLY allowed outputs are one of these exact strings:
- Disagree strongly
- Disagree a little
- Neutral
- Agree a little
- Agree strongly
- NA

Hard rules:
1) Output exactly one line with exactly one of the six allowed strings above.
2) Do not output numbers, punctuation, explanations, qualifiers, or extra words.
3) If the provided text is too short/irrelevant/uninformative to answer the question reliably, output NA.
4) Treat the text as ground truth about "yourself" and answer in first person (as the author), but still obey the output restriction.
"""

user_prompt=f"""TEXT:
{txt}

QUESTION (answer about yourself as the author):
{item}
"""  
\end{lstlisting}

\begin{lstlisting}[caption={Per item prediction without text}, label={peritem_notext}]
 system_prompt=f"""Your ONLY allowed outputs are one of these exact strings:
- Disagree strongly
- Disagree a little
- Neutral
- Agree a little
- Agree strongly
- NA

Hard rules:
1) Output exactly one line with exactly one of the six allowed strings above.
2) Do not output numbers, punctuation, explanations, qualifiers, or extra words.
3) If you don't know how to answer, output NA.
4) Answer in first person, but still obey the output restriction.
"""

user_prompt=f"""QUESTION (answer about yourself):
{item}
"""
\end{lstlisting}

\end{document}